\documentclass[11pt,a4paper]{article}
\usepackage{epsfig,graphicx}
\usepackage[hyperref]{acl2019}
\usepackage{times}
\usepackage{latexsym}
\usepackage{booktabs}
\usepackage{cleveref}
\usepackage{subcaption}

\usepackage{url}

\usepackage{todonotes}

\aclfinalcopy 

\title{Context is Key: Grammatical Error
Detection with \\ Contextual Word Representations}

\author{Samuel Bell$^{\spadesuit\diamondsuit}$ ~ Helen Yannakoudakis$^{\diamondsuit\clubsuit}$ ~ Marek Rei$^{\diamondsuit\clubsuit}$ \\
$^\spadesuit$Department of Psychology, University of Cambridge, United Kingdom \\
$^\diamondsuit$Dept. of Computer Science \& Technology, University of Cambridge, United Kingdom \\
$^\clubsuit$ALTA Institute, University of Cambridge, United Kingdom \\
{ \small \tt
sjb326@cam.ac.uk, \{helen.yannakoudakis,marek.rei\}@cl.cam.ac.uk
}
}

\date{}

\begin{document}
\maketitle

\begin{abstract}
    Grammatical error detection (GED) in non-native writing requires systems to identify a wide range of 
errors in text written by language learners. 
Error detection as a purely supervised task can be challenging, as GED datasets are limited in size and the label distributions are highly imbalanced.
 Contextualized word representations offer a possible solution, as they can efficiently capture compositional information in language and can be optimized on large amounts of unsupervised data.
In this paper, we perform a systematic comparison of ELMo, BERT and Flair embeddings \citep{Peters2017SemisupervisedST, Devlin2018,
AkbikAlanDuncanBlytheRoland2018} on a range of public GED datasets, and propose
an approach to effectively integrate such representations in current methods, achieving a new state of the art on GED.
We further analyze the strengths and weaknesses of different contextual embeddings for the task at hand, and present detailed analyses of their impact on different types of errors.
\end{abstract}

\section{Introduction}

Detecting errors in text written by language learners is a key component of pedagogical applications for language learning and assessment.
Supervised learning approaches to the task exploit
public error-annotated corpora \citep{Yannakoudakis2011AND, ng2014conll, napoles2017jfleg} that are, however, limited in size, in addition to having a biased distribution of labels: the number of correct tokens in a text far outweighs the incorrect 
\citep{leacock2014automated}.
As such, Grammatical Error Detection (GED) can be considered a
low-/mid-resource task.

The current state of the art explores error detection within a semi-supervised, multi-task learning framework, using a neural sequence labeler optimized to detect errors as well as predict their surrounding context \citep{Rei2017SemisupervisedML}. To further improve GED performance, recent work has investigated the use of artificially generated training data \citep{Rei2017,Kasewa2018}.
On the related task of grammatical error correction (GEC), \citet{Junczys-Dowmunt2018} explore transfer learning approaches to tackle the low-resource bottleneck of the task and, among others, find substantially improved performance when incorporating pre-trained word embeddings \citep{Mikolov2013Eff}, and importing network weights from a language model trained on a large unlabeled corpus.

Herein, we extend the current state of the art for error detection \citep{Rei2017SemisupervisedML} to effectively incorporate
\textit{contextual embeddings}: 
word representations that are constructed based on the context in which the words appear. 
These embeddings are typically the output of a set of hidden layers of a large
language modelling network, trained on large volumes of unlabeled and
general domain data.
As such, they are able to capture detailed information regarding language and composition from a wide range of data sources, and can help overcome resource limitations for supervised learning.

We evaluate the use of contextual embeddings in the form of Bidirectional Encoder Representations from Transformers (BERT)
\citep{Devlin2018}, embeddings from Language Models (ELMo) \citep{Peters2018}
and Flair embeddings \citep{AkbikAlanDuncanBlytheRoland2018}. 
To the best of our knowledge, this is the first evaluation of the use of contextual embeddings for the task of GED. Our contributions are fourfold:
\vspace{-0.2cm}
\begin{itemize}  
\item We present a systematic comparison of different contextualized word representations for the task of GED;
\item We describe an approach for effectively integrating contextual representations to error detection models, achieving a new state of the art on a number of public GED datasets, and make our code and models publicly available online;
\item We demonstrate that our approach has particular benefits for transferring to out-of-domain datasets, in addition to overall improvements in performance;
\item We perform a detailed analysis of the strengths and weaknesses of different contextual representations for the task of GED, presenting detailed results of their impact on different types of errors in order to guide future work.
\end{itemize}
\section{Related work}
\label{relatedwork}

In this section, we describe previous work on GED and on the related task of GEC. While
error correction systems can be used for error detection, previous work has shown that standalone error detection models can be complementary to error correction ones, and can be used to further improve performance on GEC  \cite{yannakoudakis2017neural}.

Early approaches to GED and GEC relied upon handwritten rules and error grammars (e.g.\
\citet{Foster2004}), while later work focused on supervised learning from error-annotated corpora using feature engineering approaches and often utilizing maximum entropy-based classifiers (e.g.
 \citet{chodorow2007detection, DeFelice2010}). 
A large range of work has focused on the development of systems targeting specific error types, such as 
preposition \citep{tetreault2008ups, chodorow2007detection},
article usage \citep{han2004detecting, han2006detecting}, and verb form errors
\citep{lee2008correcting}. Among others, error-type agnostic approaches have focused on generating synthetic ungrammatical data to augment the available training sets, or learning from native English datasets; for example, \newcite{Foster2009} investigate rule-based error generation methods, while \newcite{Gamon2010} trains a language model (LM) on a large, general domain corpus, from which features (e.g.\ word likelihoods)
are derived for use in error classification. 

As a distinct task, GEC has been formulated as a na\"{i}ve-bayes classification \citep{rozovskaya2013university,
rozovskaya2014illinois, Rozovskaya2016} or a monolingual (statistical or neural) machine translation (MT) problem  
(where uncorrected text is treated as the source ``language'' and the corrected text as its target counterpart) \citep{felice2014grammatical, junczys2014amu, Rozovskaya2016, yuan2016grammatical}. 

Recently, \citet{Rei2016CompositionalSL} presented the first approach towards neural GED, training a sequence labeling model based on word embeddings
processed by a bidirectional LSTM (bi-LSTM), outputting a probability
distribution over labels informed by the entire sentence as context.
This approach achieves strong results when trained and
evaluated on in-domain data, but shows weaker generalization performance on  out-of-domain data. \citet{Rei2016AttendingTC} extended this model to include character embeddings
in order to capture morphological similarities such as word endings. 
\citet{Rei2017SemisupervisedML}
subsequently added a secondary LM objective to the neural sequence labeling architecture, operating on both word and character-level embeddings. This was found to be particularly useful
for GED -- introducing an LM objective
allows the network to learn more generic features about language and composition. 
At the same time, \citet{ReiYannakoudakis2017} investigated the effectiveness of a number of
auxiliary (morpho-syntactic) training objectives for the task of GED, finding that  predicting part-of-speech tags, grammatical relations or error types as auxiliary tasks yields improvements in performance over the single-task GED objective (though not as high as when utilizing an LM objective). 

The current state of the art on GED is based on augmenting neural approaches with artificially generated training data. 
\citet{Rei2017} showed improved GED performance using the bi-LSTM sequence labeler, by generating artificial errors in two different ways: 1) learning frequent error patterns from error-annotated
corpora and applying these to error-free text; 2) using a statistical MT approach to
``translate'' correct text to its incorrect counterpart using parallel corpora. Recently, \citet{Kasewa2018} applied the latter approach using a neural MT system instead, and achieved a new state of the art on GED using the neural model of \newcite{Rei2017SemisupervisedML}. 
\section{Data} 
\label{data}

In this section, we describe the different public datasets we use to train our models. 
The First Certificate in English (FCE) dataset \citep{Yannakoudakis2011AND} is a publicly-released set of essays written by non-native learners of English taking a language assessment exam. Each essay is annotated by professional annotators with the spans of language errors committed, the types of errors, and suggested corrections.
In addition, the CoNLL 2014 shared task on GEC \citep{ng2014conll} used a dataset of English essays written by advanced undergraduate students at the National University of Singapore. Each essay is annotated by two experienced annotators and has error annotations similarly to the FCE, though using a different error taxonomy.
The Johns Hopkins University (JHU) FLuency-Extended GUG Corpus (JFLEG) dataset \citep{napoles2017jfleg} contains essays written by a range of English learners with different first languages and proficiency levels. Each essay is corrected by four annotators with native-level proficiency and annotated with fluency and grammar edits.

The 2019 Building Educational Applications (BEA) shared task on GEC \cite{Bryant2019} released two new datasets: the Cambridge English Write \& Improve (W\&I) corpus, which is a collection of texts written by learners of English of varying levels of proficiency and submitted for assessment to the Write \& Improve system  \cite{Yannakoudakis2018}, an automated online
tool for writing feedback; and the LOCNESS corpus \cite{Granger1998}, originally compiled at the Centre for English Corpus Linguistics at the University of Louvain, and comprising essays written by native English students. Both datasets are annotated for corrections by the W\&I annotators. 

In this work, we use the FCE training set as training data, and evaluate our models on the FCE test set, the CoNLL-2014 test set, the JFLEG test set, and the BEA 2019 shared task development and test sets. This setup allows us to investigate the extent to which our models and the use of contextualized representations transfer to out-of-domain data. 

We follow \citet{Rei2016CompositionalSL} and convert the span-based annotations in these datasets to binary error detection labels at the token level (i.e.\ is a token correct or incorrect). Performance is evaluated using precision, recall, and \(F_{0.5}\) at the token level. \(F_{0.5}\) places twice the weight on precision than recall: systems that incorrectly penalize correct language can have a much more negative impact to language learning compared to systems that miss to detect some errors \citep{ng2014conll}. 
We note that performance on the BEA shared task test set is conducted using the official evaluation tool in CodaLab. 

We also perform detailed analyses in order to evaluate the performance of our models per error type. As the datasets above either have their own error type taxonomy or they are not annotated with error types at all, we follow the 2019 BEA shared task and induce error types for all datasets automatically using the ERRor ANnotation Toolkit (ERRANT) \citep{Bryant2017Automatic}. ERRANT automatically annotates parallel uncorrected and corrected sentences with error types using a universal error taxonomy and hence allowing for comparisons across datasets. The system uses distance-based alignment followed by rule-based error categorization. An error type is hence assigned to every incorrect token in each dataset, with the exception of the BEA 2019 shared task test set, for which the corrected versions are not yet publicly available.

\begin{figure*}[t]
    \centering
    \begin{subfigure}[b]{\textwidth}
        \centering
        \includegraphics[width=0.65\textwidth]{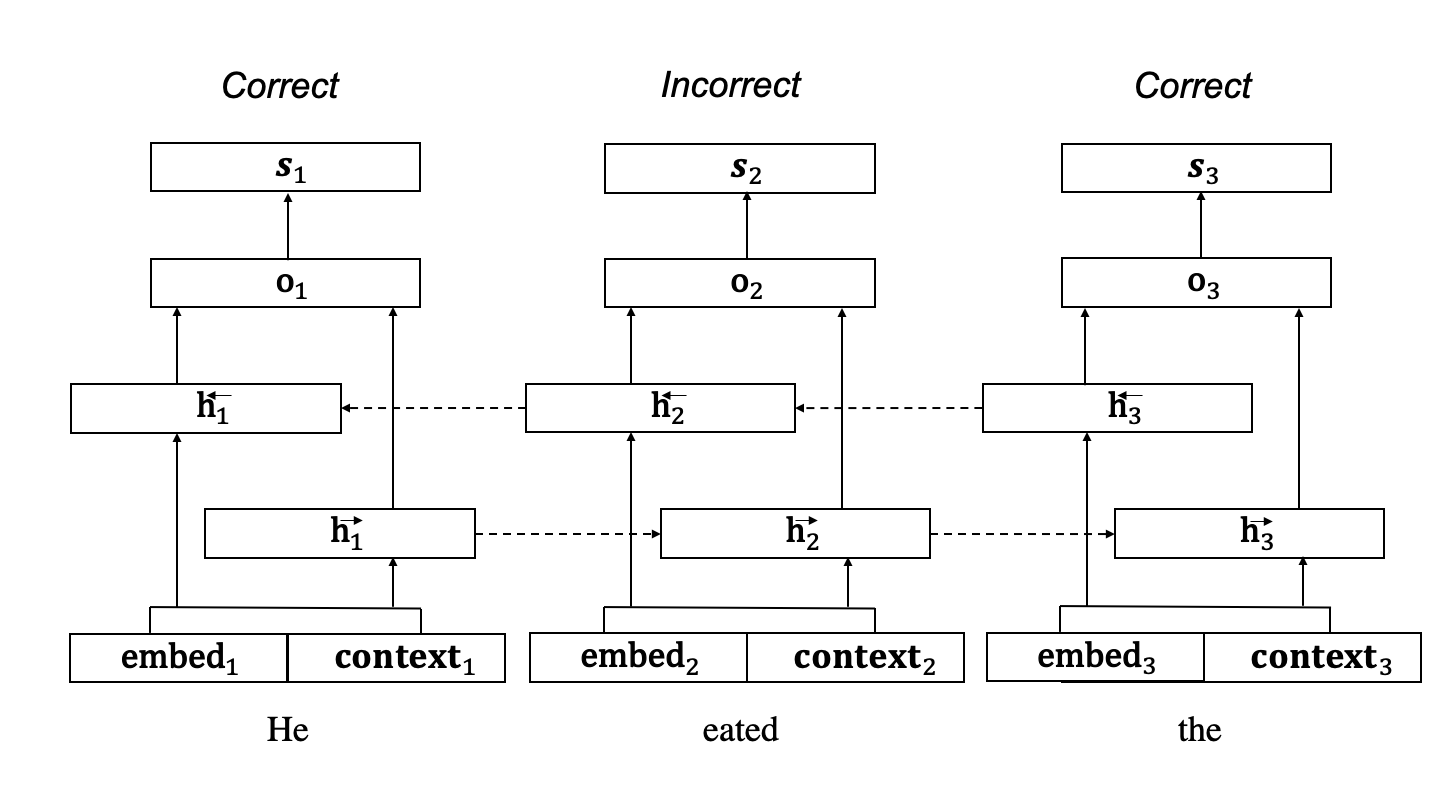}
        \caption{}
        \label{fig:diagram-input} 
    \end{subfigure}
    \begin{subfigure}[b]{\textwidth}
        \centering
        \includegraphics[width=0.65\textwidth]{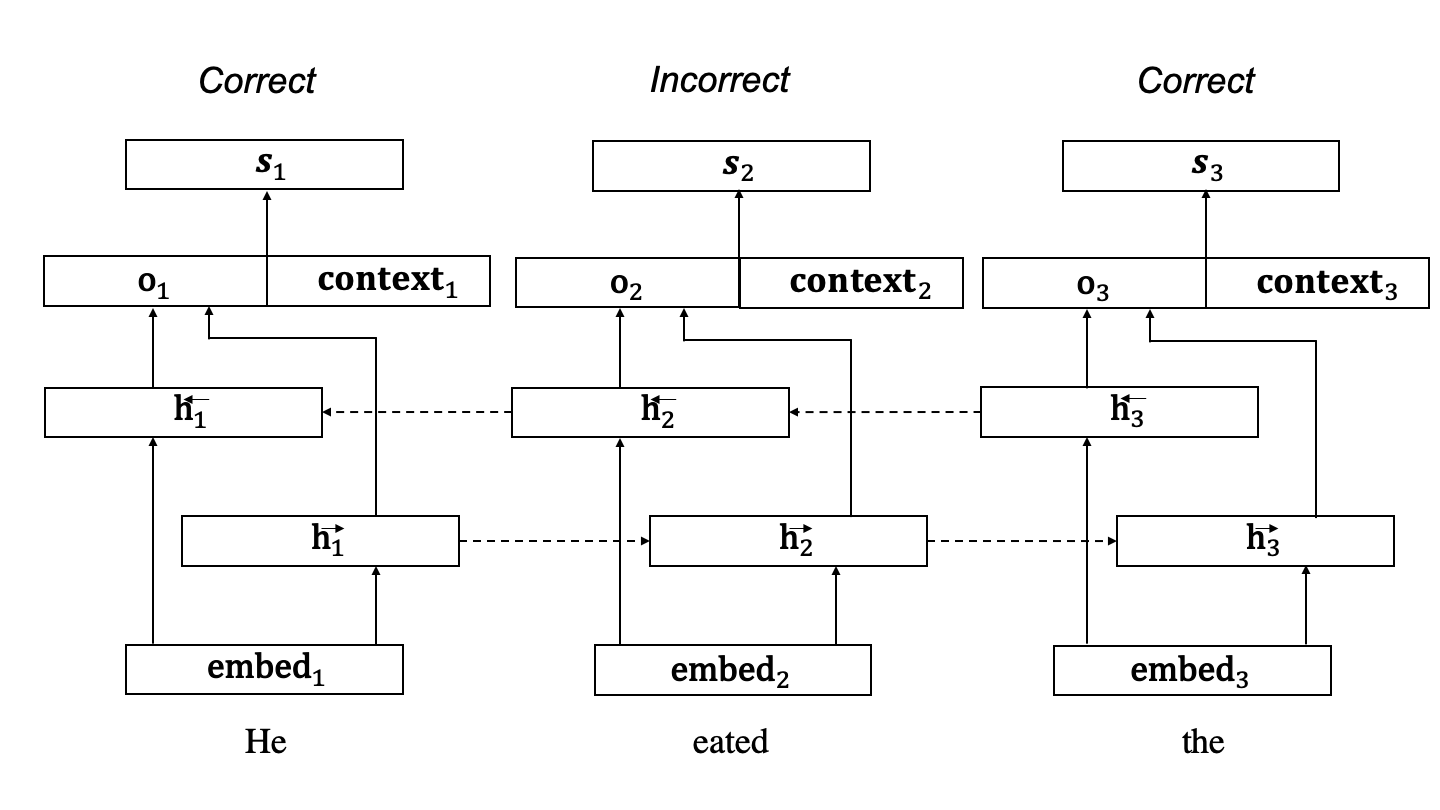}
        \caption{}
        \label{fig:diagram-output} 
    \end{subfigure}
    \caption{Simplified bi-LSTM sequence labeler with: \textbf{(a)} contextual embeddings (\(\mathbf{context}\))
concatenated to the input word embeddings (\(\mathbf{embed}\)), before being passed
through the bi-LSTM (\(\mathbf{h}\)); \textbf{(b)} contextual embeddings (\(\mathbf{context}\)) concatenated to the LSTM output (\(\mathbf{o}\)) before being passed through a softmax layer (\(\mathbf{s}\)).}

\end{figure*}
\section{Error detection model}
\label{approach}

In this section, we extend the current state of the art (neural) architecture for GED
\cite{Rei2017SemisupervisedML}, which we use as our baseline system.
This model is a bi-LSTM sequence labeler over token embeddings where, for each
token, the model is trained to output a probability distribution over binary
correct/incorrect labels using a softmax layer (i.e.\ predicting whether a
token is correct or incorrect in context). The model is additionally trained
with a secondary bidirectional LM objective, predicting
the surrounding context of the target token in the sequence. Specifically, the model uses a forward LM to predict the next token in the sequence, and a backward LM to predict the previous token. \citet{Rei2017SemisupervisedML} also makes use of a character-level bi-LSTM, as opposed to solely conditioning on tokens, in order to benefit from sub-word morphological units, of particular use in the case of unknown or incorrectly spelled words.
The outputs of the character-level LSTMs are concatenated to the word
embeddings and given as input to the word-level bi-LSTM.

The model learns $300$-dimensional word embeddings, initialized
with pre-trained Google News embeddings \citep{Mikolov2013Eff},\footnote{\url{https://code.google.com/archive/p/word2vec/}} and $100$-dimensional character embeddings. The hidden states of the word- and character-level LSTMs are also of $300$ and $100$ dimensions respectively. The outputs of each LSTM are passed through a $50$-dimensional hidden layer with a \(tanh\) activation function. Dropout is applied to the inputs and outputs of each LSTM with a probability of $0.5$. The model is trained with a cross-entropy
loss function for the error detection objective that minimizes the negative log
probability of the gold label. As the model is also trained with a
secondary LM objective, a second bipartite cross-entropy loss function is used,
minimizing the negative log probability of the next word in the sequence for
the forward LM, and the previous word for the backward LM. A
hyperparameter $\gamma=0.1$ weights the combination of the two loss functions, assigning more importance to the main task of error detection over the auxiliary task of language modelling.
Optimization is performed with the AdaDelta
optimizer \citep{zeiler2012adadelta}, using an initial learning rate of $1.0$,
and batches of $32$ sentences. Training is terminated
when validation performance does not improve for $7$ epochs. 

In this work, we extend the above model by incorporating contextualized word embeddings, produced by three different approaches (BERT, ELMo and Flair; each described in more detail in
\Cref{sec:contextual-embeddings}).
Specifically, we concatenate the contextual embeddings either to
the input word embeddings before being passed through the word-level bi-LSTM
(\Cref{fig:diagram-input}),
or to the bi-LSTM's output (\Cref{fig:diagram-output}).
\citet{Peters2018} find that the best point to integrate ELMo vectors varies by task and, as such, we continue that line of analysis here. 

We make a TensorFlow \citep{abadi2016tensorflow} implementation of our code and
models publicly available
online.\footnote{\url{https://github.com/samueljamesbell/sequence-labeler}} 

\begin{table*}[t]
    \centering
        \small
\begin{tabular}{lrrrrrrrrr}
\toprule
                                    & \multicolumn{3}{c}{CoNLL Test 1}                       & \multicolumn{3}{c}{CoNLL Test 2}                       & \multicolumn{3}{c}{FCE test} \\
                                    &     \(P\)         &     \(R\)         & \(F_{0.5}\)       &     \(P\)         &     \(R\)         & \(F_{0.5}\)       & \(P\)             &     \(R\)         & \(F_{0.5}\)       \\
\midrule                                                                                                                                                                                                                
\citet{Rei2017SemisupervisedML}     & 17.68             & 19.07             & 17.86             & 27.6              & 21.18             & 25.88             & 58.88             & 28.92             & 48.48             \\
\citet{Rei2017}                     & 23.28             & 18.01             & 21.87             & 35.28             & 19.42             & 30.13             & 60.67             & 28.08             & 49.11             \\
\citet{Kasewa2018}                  & -                 & -                 & 28.3              & -                 & -                 & 35.5              & -                 & -                 & 55.6              \\
\midrule                                                                                                                                                                                                                
Baseline                            & 20.82             & 16.31             & 19.73             & 31.91             & 17.81             &         27.55     & 46.55             & 30.58             & 42.15             \\
\midrule                                                                                                                                                                                                                
Flair                               & 29.53             & 17.11             & 25.79             & 44.12             & 18.22             &         34.35     & 58.36             & 31.72             & 49.97             \\
ELMo                                & 30.83             & 23.90             & 29.14             & 46.66             & 25.77             &         40.15     & 58.50             & 38.01             & 52.81             \\
BERT base                           & 37.62             & 29.65             & 35.70             & \textbf{53.52}    & 30.05             & \textbf{46.29}    & 64.96             & \textbf{38.89}    & \textbf{57.28}    \\
BERT large                          & \textbf{38.04}    & \textbf{33.12}    & \textbf{36.94}    & 51.40             & \textbf{31.89}    &         45.80     & \textbf{64.51}    & 38.79             & 56.96             \\
\bottomrule
\end{tabular}

    \caption{Error detection precision, recall, and \(F_{0.5}\) on the
    FCE and CoNLL-2014 test sets: test 1 and test 2 refer to the two different CoNLL annotators. `Baseline' refers to our own re-training of the model by \citet{Rei2017SemisupervisedML}.}\label{table:fce-conll-results}
\end{table*}

\begin{table*}[t]
    \centering
        \small
\begin{tabular}{lrrrrrrrrr}
\toprule
                & \multicolumn{3}{c}{JFLEG Test}                            & \multicolumn{3}{c}{Shared Task Dev}                           & \multicolumn{3}{c}{Shared Task Test} \\
                &             \(P\) &             \(R\) &       \(F_{0.5}\) &             \(P\) &             \(R\) &           \(F_{0.5}\) & \(P\)             &     \(R\)         &     \(F_{0.5}\)  \\
\midrule                                                                                                                                                                                               
Baseline        & 72.84             & 22.83             &           50.65   & 31.31             & 21.18             &      28.58            & 40.05             & 34.99             & 38.93            \\
\midrule                                                                                                                                                                                               
Flair           & 75.65             & 25.26             &           54.08   & 41.80             & 24.10             &      36.45            & 53.40             & 39.84             & 50.00            \\
ELMo            & 74.95             & 31.21             &           58.54   & 47.90             & 30.41             &      42.96            & 58.72             & 47.79             & 56.15            \\
BERT base       & \textbf{79.51}    & 32.94             & \textbf{61.98}    & \textbf{53.31}    & 35.65             &      \textbf{48.50}   & \textbf{66.47}    & \textbf{54.11}    & \textbf{63.57}   \\
BERT large      & 76.47             & \textbf{34.52}    &           61.52   & 51.54             & \textbf{36.90}    &      47.75            & 63.35             & 54.10             & 61.26            \\
\bottomrule
\end{tabular}

    \caption{Error detection precision, recall, and \(F_{0.5}\) on the
    JFLEG test set and BEA 2019 GEC Shared Task development and test
sets.}\label{table:jfleg-bea-dev-results}
\end{table*}

\section{Contextualized embeddings}

\label{sec:contextual-embeddings}

Three types of contextual embeddings are considered in this work: BERT, ELMo
and Flair embeddings \citep{Peters2017SemisupervisedST,
Devlin2018,AkbikAlanDuncanBlytheRoland2018}. In each case, we use
the publicly-available pre-trained models released by the authors.

BERT embeddings are extracted from the highest layers of a transformer
architecture trained with a masked LM objective: rather than predicting the next or previous word in a sequence, a percentage of input tokens are masked and then the network learns to predict the masked tokens. BERT is also trained with a second objective predicting whether one sentence
directly follows another, given two input sentences. BERT pre-trained
embeddings are available in two variants: \textit{base} embeddings, which are the
concatenation of the four highest $768$-dimension hidden layers, yielding a
$3,072$-dimension embedding; \textit{large} embeddings, which are the concatenation of
the four highest $1024$-dimension hidden layers, yielding a $4,096$-dimension
embedding \citep{Devlin2018}. BERT embeddings are trained on the BooksCorpus
($0.8$ billion words) of books written by unpublished authors \citep{Zhu2015}
and English Wikipedia ($2.5$ billion words).

ELMo embeddings are a weighted element-wise sum of the outputs of three-layered stacked bi-LSTM LMs, trained to predict both the next and previous token in the
sequence. Using a task-specific scalar per layer, the outputs of the three LSTMs are reduced to a single $1,024$-dimension embedding \citep{Peters2018}. This task-specific weighting is learned by our sequence labeler during
training. ELMo is trained on the One Billion Word Benchmark corpus ($0.8$ billion words), composed primarily of online news articles \citep{Chelba2014}.

Flair embeddings are the concatenated output of a single (i.e.\ unstacked)
character-level bi-LSTM. We use the concatenation of both the $2,048$-dimension
``news-forward'' and ``news-backward'' embeddings, each produced by a forward and backward bi-LSTM respectively, and both
trained on the One Billion Word Benchmark \citep{chelba2014one}. This yields a $4,096$-dimensional embedding
\citep{AkbikAlanDuncanBlytheRoland2018}.

\begin{table*}[t]
    \centering
        \small
\begin{tabular}{llrrrrr}
\toprule
      & & Shared Task Dev & CoNLL Test 1 & CoNLL Test 2 & FCE test & JFLEG test \\
\midrule
Flair       & Input     &      36.45 &         25.79 &         34.35 &           49.97 &           54.08 \\
            & Output    &      33.47 &         24.52 &         33.18 &           48.50 &           52.10 \\
\midrule
ELMo        & Input     &      42.96 &         29.14 &         40.15 &           52.81 &           58.54 \\
            & Output    &      37.33 &         27.33 &         38.10 &           52.99 &           54.86 \\
\midrule
BERT base   & Input     &      \textbf{48.50} &         35.70 &         46.29 & \textbf{57.28} &           \textbf{61.98} \\
            & Output    &      46.33 &         37.04 &         46.50 &           55.32 &           60.97 \\
\midrule
BERT large  & Input     &      47.75 &         36.94 &         45.80 &           56.96 &           61.52 \\
            & Output    &      46.72 &  \textbf{39.07} & \textbf{46.96} &           55.10 &           60.56 \\
\bottomrule
\end{tabular}

    \caption{Error detection \(F_{0.5}\) of different embedding integration strategies (`input' vs.\ `output') per model on all datasets.}\label{table:input-output-results}
\end{table*}

\begin{figure*}[t]
\centering
\includegraphics[width=\textwidth]{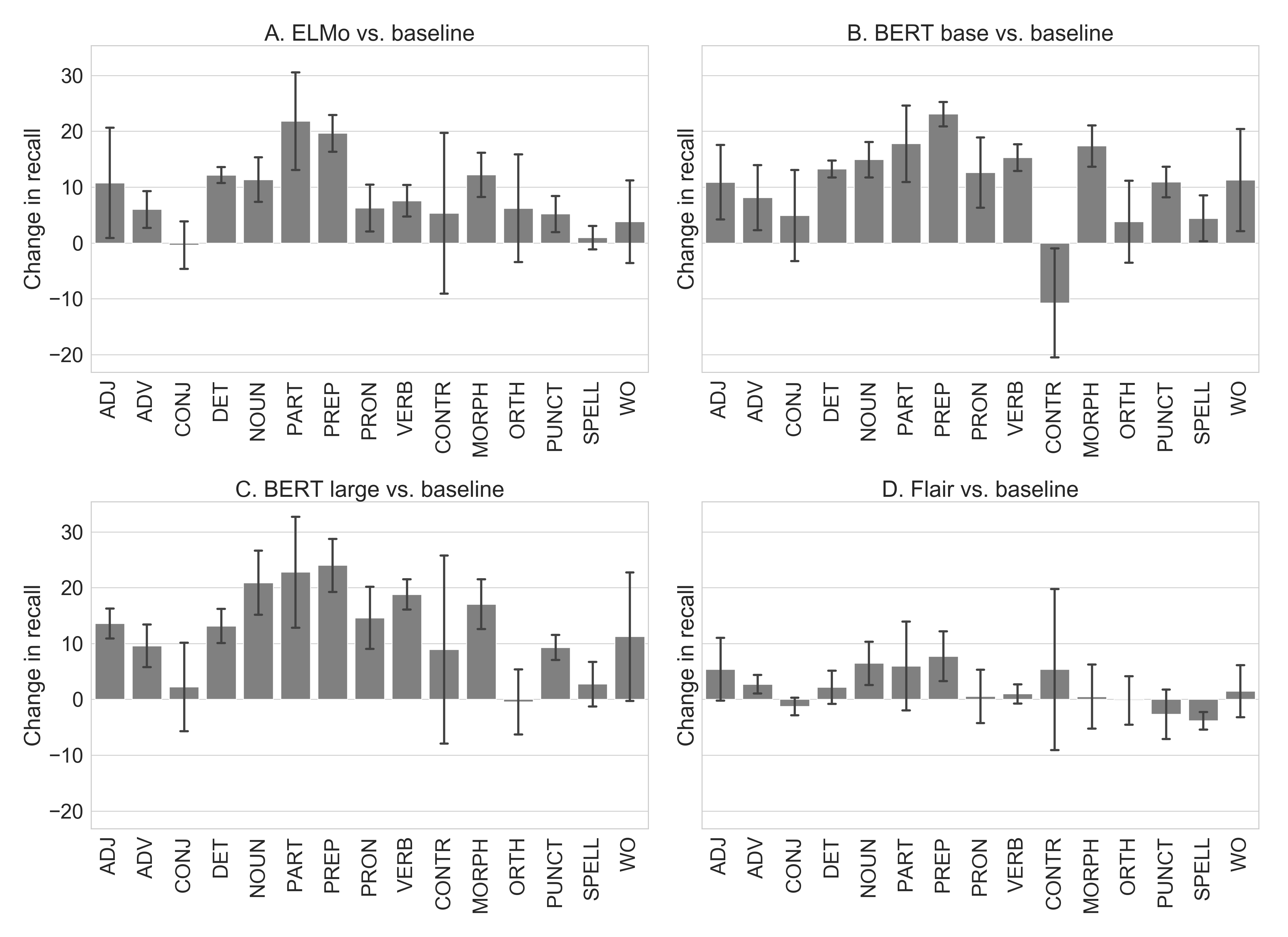}
\caption{Mean change in recall of ERRANT-induced POS-based error types over all
datasets when adding contextual embeddings at the input level, vs.\ our
baseline without contextual embeddings. \textbf{A:} ELMo. \textbf{B:} BERT
base. \textbf{C:} BERT large. \textbf{D:} Flair.}
\label{fig:recall-by-error-token-vs-baseline} 
\end{figure*}

\begin{figure*}[t]
\centering
\includegraphics[width=\textwidth]{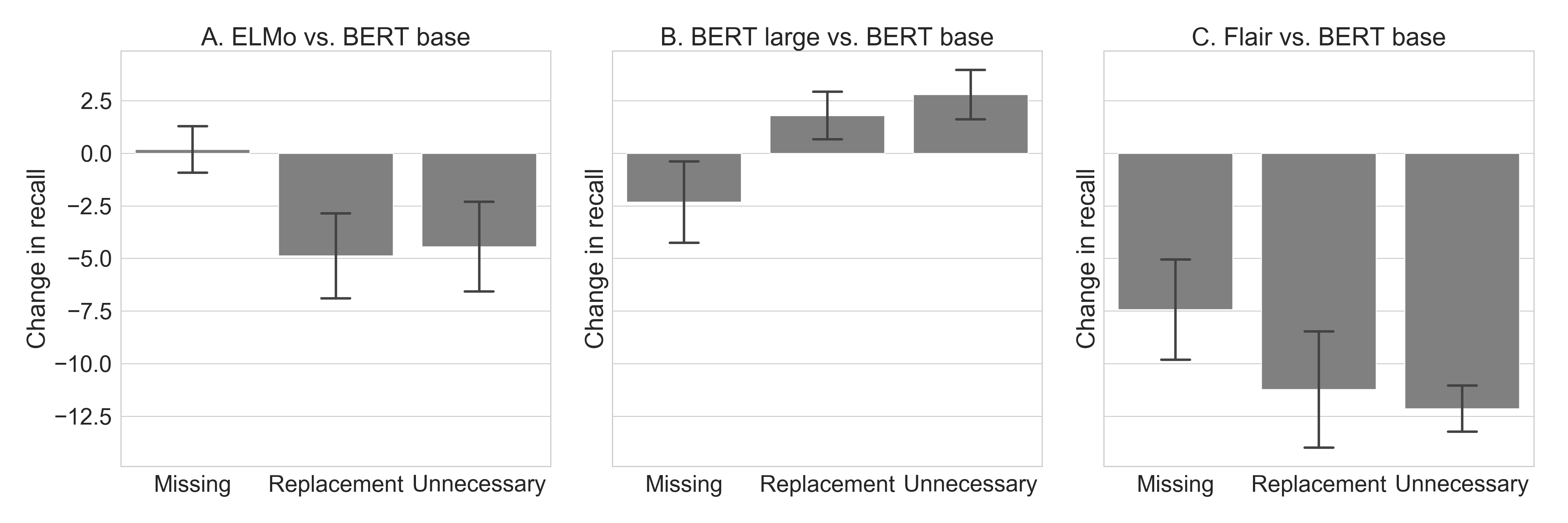}
\caption{Mean change in recall of ERRANT-induced edit operation error types
across all datasets when adding contextual embeddings at the input level, vs.\
the model using BERT base. \textbf{A:} ELMo. \textbf{B:} BERT large. \textbf{C:} Flair.}
\label{fig:recall-by-error-operation-vs-bert} 
\end{figure*}

\section{Results}


\Cref{table:fce-conll-results} and \Cref{table:jfleg-bea-dev-results} present
the results of integrating different contextual embeddings with the current
state-of-the-art model described by
\citet{Rei2017SemisupervisedML}.\footnote{We include the results reported by
\citet{Rei2017SemisupervisedML} along with our re-trained baseline. We note that the differences in performance are due to a re-processing of the data in order to align parallel original--corrected sentences and derive fine-grained error type labels for later analyses (see Section \ref{errsect}).}
The experiments in this section are based on models with contextual representations concatenated to the word embeddings; Section \ref{strategies} includes a more detailed investigation of different integration points.
For comparison, we also report the results of \citet{Rei2017}
and \citet{Kasewa2018}, who improve error detection performance by additionally
augmenting \newcite{Rei2017SemisupervisedML}'s model with artificial training
data. 

The experiments demonstrate a substantial improvement in precision, recall and
\(F_{0.5}\) for every model incorporating contextual embeddings, across every dataset considered. 
On the FCE test set, even our lowest performing model (Flair, \(F_{0.5}=49.97\)) outperforms the baseline  (\(F_{0.5}=42.15\)), with a relative improvement of $18.55\%$. Our best performing model (BERT base, \(F_{0.5}=57.28\)) outperforms the baseline by a relative $35.9\%$. This is also the new state-of-the-art result on the FCE test set, without using additional manually-annotated training data.

The best performance on the CoNLL-2014 test set is achieved by BERT large ($F_{0.5}=36.94$) and BERT base ($F_{0.5}=46.29$) for the first and second annotator
respectively.
These scores show more than $30\%$ relative improvement over the previous best results by \citet{Kasewa2018}, even without using additional artificial training data, for both annotators.
On the JFLEG test set (\Cref{table:jfleg-bea-dev-results}),
and both the BEA 2019 GEC Shared Task development and test sets,
BERT base yields the highest performance. 
The improvement on the BEA shared task datasets is particularly large, with BERT base achieving $69.70\%$ relative improvement on the development set and $63.30\%$ relative improvement on the test set, compared to the baseline model.

These experiments demonstrate that contextual embeddings provide a very beneficial addition for GED systems, achieving a new state of the art across all datasets.
Learning to compose language representations on large unsupervised datasets allows the models to  access a wider range of useful information. While our error detection models are optimized on the FCE training set, we observe particularly large improvements on the CoNLL-2014 and BEA shared task datasets, indicating that contextual embeddings allow the models to generalize better and capture errors in out-of-domain texts. Overall, we found BERT base to provide the highest improvements across all datasets. The slightly lower performance of BERT large could be attributed to the larger embedding sizes requiring more parameters to be optimized on our limited GED dataset. 

\subsection{Integration method}
\label{strategies}
We performed additional experiments to investigate the optimal method for integrating contextual embeddings into the error detection model.
The embeddings are either concatenated to the standard word embeddings at the
input level (reported in our results as `input'), or to the output of the
word-level bi-LSTM (reported as `output'). In all experiments, contextual
embeddings are not fine-tuned.

\Cref{table:input-output-results} compares the \(F_{0.5}\)
of these two strategies for each model, across all the datasets. 
We observe that, although performance varies across datasets and models, integration by concatenation to the word embeddings yields the best results across the majority of datasets for all models (BERT: 3/5 datasets; ELMo: 4/5 datasets; Flair: 5/5 datasets).
The lower integration point allows the model to learn more levels of task-specific transformations on top of the contextual representations, leading to an overall better performance.

\subsection{Error type performance}
\label{errsect}

Using ERRANT \citep{Bryant2017Automatic}, we analyze the performance on different types of errors and specifically focus on two error type taxonomies: one that uses part-of-speech (POS) based error types (i.e.\ the type is based on the POS tag of the incorrect token), and another based on edit operations (i.e.\ is it a missing token, an unnecessary token, or a replace token error).
This allows us to yield insights into how different types of contextual
embeddings and the data in which they were trained may impact performance on
specific errors. 
Since identifying type-specific false positives is not possible in this setting, we follow \citet{ng2014conll} and report recall on each error type.

\Cref{fig:recall-by-error-token-vs-baseline} presents the performance on POS-based error types,
showing the change in error type recall of each contextual embedding model compared to our baseline, averaged over all datasets. 
While all models yield an improvement in aggregate performance metrics (\(P, R,
F_{0.5}\)), when broken down by POS-based error type, some trends emerge. BERT
base, BERT large and ELMo each show strong improvements in recall of errors
relating to nouns, particles, prepositions and morphology. In comparison, relatively weak improvements are achieved for errors of conjugation, orthography and
spelling. 
As such words/errors are less likely to occur frequently in general-purpose
corpora of English (i.e.\ spelling mistakes are less frequent in news articles compared to learner essays), contextual embeddings trained on such corpora are also less helpful for these error types.

We also note the sharp decline in recall of the BERT base model on contraction
errors. This error type occurs quite infrequently (see Figure $1$ in the Appendix)
and we do not observe this issue with BERT large.

Compared to BERT and ELMo, Flair offers very little improvement
on POS-based error type recall or even actively
degrades recall (e.g.\ conjugation, punctuation or spelling errors). While the purely character-based representations of Flair could potentially offer more flexibility in the model, these results suggest that the limited vocabulary of our learner data may be better captured with word-level approaches. 

\Cref{fig:recall-by-error-operation-vs-bert}
presents the differences between models when looking at error types based on the necessary edit operation: missing token, replace token or unnecessary token.
While BERT improves overall performance compared to ELMo, this improvement appears to be limited to replacement and unnecessary error types. On missing word errors, BERT base performs comparably to ELMo and BERT large even decreases performance. We discuss the possible reasons for this in Section \ref{sec:discussion}.

We include the full results table for different error types in the Appendix (Table 1, Table 2).
Our analysis shows that focusing more on punctuation, determiner and preposition errors might be the most beneficial avenue for improving the overall performance on GED.
For example, punctuation errors are the third most common error type, but even with contextual embeddings the models only achieve $27.7\%$ recall across all datasets.

Overall, our results suggest that, while contextual embeddings always improve
aggregate performance on GED, error type specific properties of models should also be considered.
\section{Discussion}
\label{sec:discussion}

The previous section has demonstrated consistent improvement in precision, recall and \(F_{0.5}\) across a range of GED datasets, including many examples of transfer to different domains, irrespective of the choice of contextual embedding.
Contextual embeddings bring the possibility of leveraging information learned in an unsupervised manner from high volumes of unlabeled data, by large and complex models, trained for substantial periods of time. For these reasons, they are a particularly appropriate addition to a low-resource task such as GED.
While each choice of contextual embedding yields improved performance, BERT
base  and BERT large consistently outperform ELMo and Flair embeddings. Here,
we suggest that details of the BERT architecture and training process may be
responsible for its specific performance profile.

One difference between contextual embedding models is the choice of training corpora. While both
ELMo and Flair embeddings are trained using the One Billion Word Benchmark,
BERT is trained on the BooksCorpus, and
English Wikipedia. It is likely that the usage and distribution of English
varies across these corpora, yielding different results on downstream tasks. We
might expect a corpus of books to exhibit a greater variance in writing style,
audience, and even writer ability than a corpus of news articles, perhaps
resulting in more useful contextual embeddings for GED. However, 
in contrast to the $0.8$ billion tokens of training data available to
ELMo and Flair, BERT's combination of BooksCorpus and English Wikipedia
provides $3.3$ billion tokens of training data. The increased volume of training
data may alone suffice to explain BERT's comparatively strong performance.

Another difference is that BERT is not trained with a bi-directional LM
objective. In contrast to ELMo and Flair, BERT is trained with a masked LM objective, such that it
must predict the original tokens when presented with a sentence with any number
of tokens masked.
This training objective always provides the model access to the correct number of tokens in each sentence, which means it never needs to consider possible missing tokens.
This could explain the decreased improvements on the ``missing'' error types compared to other
error operations (\Cref{fig:recall-by-error-operation-vs-bert}), and future work could experiment with integrating missing tokens directly into the BERT training objective. 

At this point, we note that while the number of parameters, and
the dimensionality of the resulting representations vary by model, 
extensive ablation studies
\citep{Devlin2018, Peters2018} indicate only small decreases in performance
with decreasing layers or dimensionality. Future research may contrast the
models considered here with those of a paired number of parameters but with
randomly-initialized contextual embeddings.
However, as 
contextual embeddings enable the integration of information
captured via unsupervised learning on large general-purpose corpora, we
expect that embeddings without this information (i.e.\ randomly-initialized) 
would not yield the degree of improvement detailed herein. 

\section{Conclusion}

We have experimentally demonstrated that using contextual embeddings
substantially improves GED performance, achieving a new state of the art on a number of public datasets. We have shown that a sequence labeling architecture augmenting the input word embeddings with the BERT family (base or large) of contextual embeddings produces, overall, the best performance across datasets.
In addition to improving overall performance, contextual embeddings proved to be particularly useful for improving on out-of-domain datasets. 

We have also performed a detailed analysis of the strengths and weaknesses of
the use of different contextual embeddings on detecting specific types of
errors. We aim for the analyses presented here to facilitate future work in GED
and in improving such systems further. Future work can also be directed towards
investigating alternative approaches to integrating contextualized
representations and fine-tuning such representations for GED.

\section*{Acknowledgments}
We thank the anonymous reviewers for their valuable feedback. Marek Rei and Helen Yannakoudakis were supported by Cambridge Assessment, University of Cambridge.

\bibliography{acl2019}
\bibliographystyle{acl_natbib}

\newpage
\onecolumn
\setcounter{figure}{0}
\setcounter{table}{0}
\appendix
\section{Supplementary figures}

\begin{figure}[h]
\centering
\includegraphics[width=0.5\columnwidth]{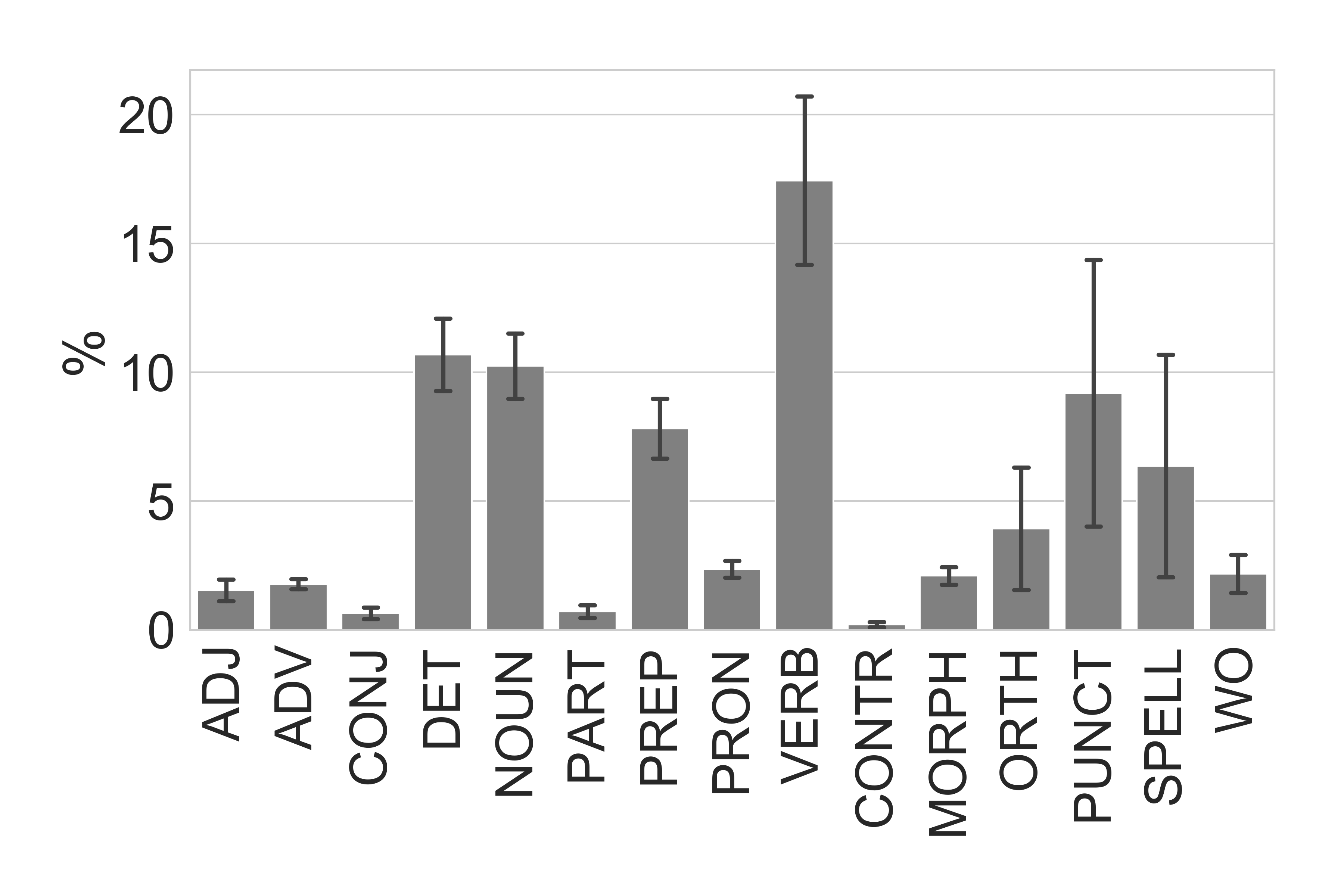}
\caption{Mean proportion of ERRANT-induced POS-based error types across datasets (FCE test set, CoNLL 2014 test set 1 (both annotators), BEA 2019 GEC Shared Task development set and JFLEG test set). Error bars show the standard deviation.}\label{fig:distribution-error-token} 
\end{figure}

\begin{figure}[h]
\centering
\includegraphics[width=0.5\columnwidth]{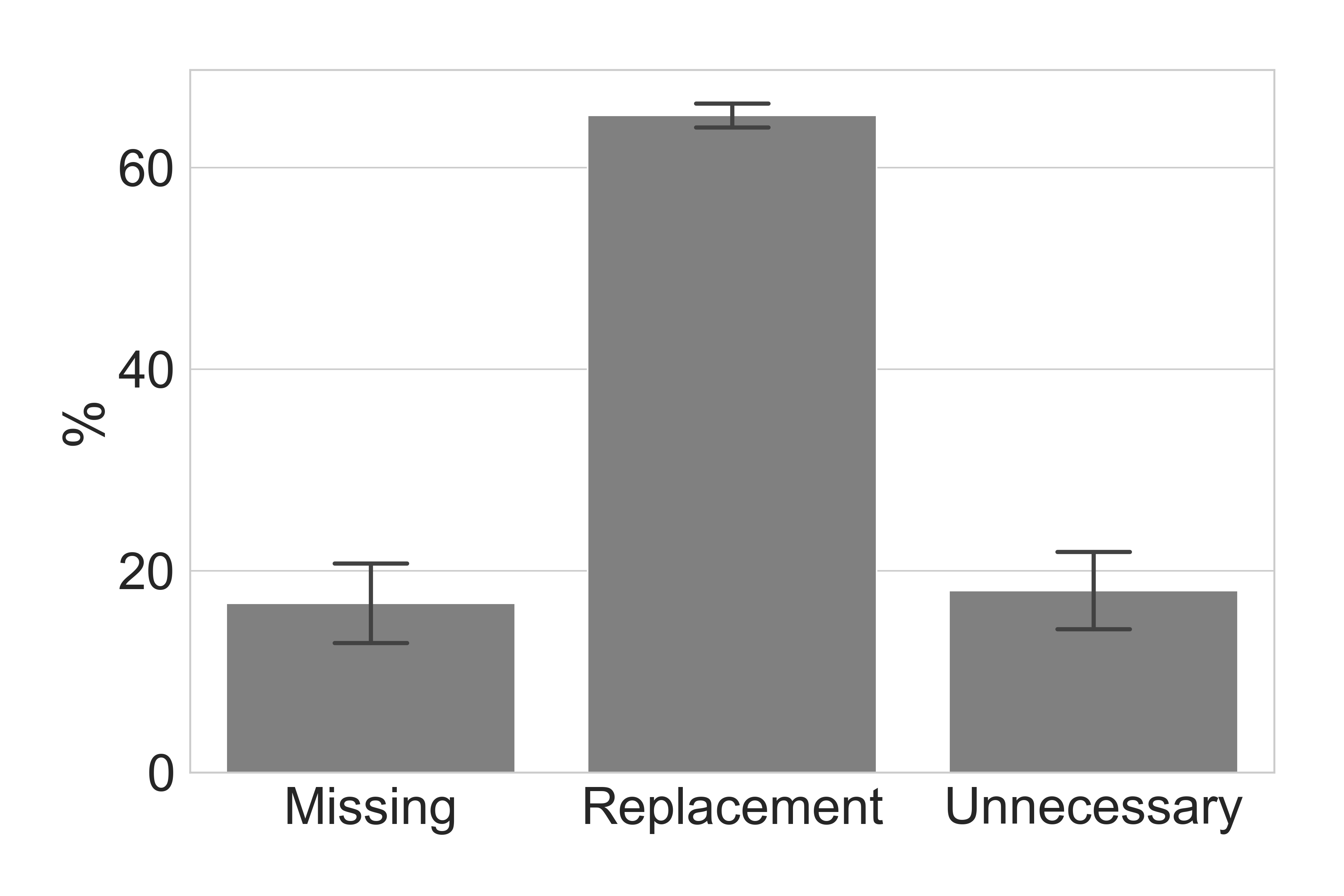}
\caption{Mean proportion of ERRANT-induced edit operation error types across all datasets (FCE test set, CoNLL 2014 test set (both annotators), BEA 2019 GEC Shared Task development set and JFLEG test set).  Error bars show the standard deviation.}
\label{fig:distribution-error-operation} 
\end{figure}

\newpage

\section{Supplementary tables}

\begin{table*}[h]
    \centering
        \begin{tabular}{lrrrrr|r}
\toprule
    {}  & \multicolumn{5}{c}{\(R\)} & \\
    & Baseline &  BERT base & BERT large &  ELMo & Flair &  Frequency \\
\midrule
ADJ         &    16.35 & 25.47 &      29.49 & 24.13 & 18.77 &         373 \\
ADV         &     8.18 & 16.62 &      17.39 & 13.81 & 11.25 &         391 \\
CONJ        &     9.63 & 15.56 &      12.59 &  8.89 &  8.15 &         135 \\
CONTR       &    18.60 & 11.63 &      20.93 & 25.58 & 25.58 &          43 \\
DET         &    24.78 & 38.11 &      38.02 & 37.28 & 27.30 &        2304 \\
MORPH       &    36.34 & 53.12 &      52.69 & 47.53 & 35.48 &         465 \\
NOUN        &    26.28 & 40.53 &      45.97 & 36.26 & 31.45 &        2245 \\
ORTH        &    29.40 & 28.96 &      28.20 & 30.49 & 27.76 &         915 \\
OTHER       &    18.77 & 29.96 &      31.06 & 27.04 & 20.69 &        4800 \\
PART        &     9.62 & 28.85 &      31.41 & 31.41 & 16.67 &         156 \\
PREP        &    10.70 & 34.71 &      35.25 & 31.40 & 20.42 &        1841 \\
PRON        &    12.20 & 25.98 &      28.35 & 19.69 & 13.78 &         508 \\
PUNCT       &    17.21 & 27.72 &      26.32 & 22.80 & 15.97 &        2504 \\
SPELL       &    88.33 & 92.88 &      91.48 & 89.06 & 84.88 &        1362 \\
VERB        &    18.63 & 33.75 &      36.80 & 25.40 & 19.32 &        3929 \\
WO          &    13.52 & 23.16 &      22.13 & 19.47 & 14.75 &         488 \\
\bottomrule
\end{tabular}

    \caption{Overall recall of each model over all datasets broken out by ERRANT-induced POS-based error type, with frequency of occurrence of each error type.}\label{table:appendix-recall-by-error-token}
\end{table*}

\begin{table*}[h]
    \centering
        \begin{tabular}{lrrrrr|r}
\toprule
    {} & \multicolumn{5}{c}{\(R\)} & \\
    & Baseline & BERT base & BERT large & ELMo & Flair &    Frequency\\
\midrule
Missing               &    19.00 & 28.98 &      26.83 & 29.17 & 22.48 &        3816 \\
Replacement               &    27.02 & 39.94 &      41.60 & 35.02 & 28.63 &       14839 \\
Unnecessary               &    15.38 & 29.07 &      31.81 & 24.50 & 16.72 &        3804 \\
\bottomrule
\end{tabular}

    \caption{Overall recall of each model over all datasets broken out by ERRANT-induced edit operation error type, with frequency of occurrence of each error type.}\label{table:appendix-recall-by-error-operation}
\end{table*}

\end{document}